\documentclass[11pt,a4paper]{article}
\usepackage[table,x11names]{xcolor}
\usepackage[hyperref]{acl2021}
\usepackage{times}
\usepackage{latexsym}
\usepackage{graphics}
\usepackage{graphicx}
\usepackage{booktabs}
\usepackage{comment}

\newcommand{\bestres}{\cellcolor{green}}
\usepackage{microtype}

\aclfinalcopy 

\title{Computer-Assisted Construct Classification of Organizational Performance concerning Different Stakeholder Groups}

\author{\shortstack{Seethalakshmi Gopalakrishnan \\
  University of North Carolina at Charlotte \\ 
  \texttt{sgopala4@uncc.edu}} 
 &
  \shortstack{Gus Hahn-Powell \\
  University of Arizona \\ 
  \texttt{hahnpowell@arizona.edu}} \\
 \\
  \shortstack{Victor Zitian Chen \\
  University of North Carolina at Charlotte \\ 
  \texttt{zchen23@uncc.edu}}
 &
  \shortstack{Bharadwaj Tirunagaru\\
  University of North Carolina at Charlotte \\ 
  \texttt{btirunag@uncc.edu}} \\

  }

\date{}

\begin{document}
\maketitle
\begin{abstract}
The number of research articles in business and management has dramatically increased along with terminology, constructs(i.e., conceptual terms that describe a distinct entity), and measures. Proper classification of organizational performance constructs from research articles plays an important role in categorizing the literature and understanding to whom its research implications may be relevant. In this work, we classify \textit{constructs} (i.e., concepts and terminology used to capture different aspects of organizational performance) in 125 research articles collected from ISI Web of Science database into a three-level categorization: (a) performance and non-performance categories (Level 0); (b) for performance constructs, stakeholder group-level of performance concerning investors, customers, employees, and the society (community and natural environment) (Level 1); and (c) for each stakeholder group-level, subcategories of different ways of measurement (Level 2). 
We observed that increasing contextual information with features extracted from surrounding sentences and external references 
improves classification of disaggregate-level labels, given limited training data. Our research has implications for computer-assisted construct identification and classification - an essential step for research synthesis.

\end{abstract}

\section{Introduction}
Organizational performance is defined as the extent to which an organization is effective in creating and delivering well-being for its stakeholders (e.g., investors, customers, employees, and the community). According to ISI Web of Science, there are thousands of research articles related to organizational performance management and more being published each day. A keyword search of ``organizational performance'' yields more than 28,000 documents in the year 2021 alone. A persistent challenge for a timely synthesis of this literature is the diversity and proliferation of constructs (i.e., concepts and terminologies) used to capture different aspects of organizational performance. 
\citet{kang2020natural} summarizes the toolkits and procedural steps that are critical for employing NLP in management research. This work reviews the NLP methodologies that are used in research by dividing them into four parts: text preprocessing, text representation, model training, and model evaluation. 
In this research, we explore novel ways of   classifying organizational performance constructs using various natural language processing (NLP) and machine learning (ML) techniques. 

A collection of  research papers are collected from quality refereed journals, and the hypotheses which have the organizational performance constructs are detected and extracted through a machine reading algorithm developed by \citet{chen2021machine} . From these hypotheses, both the organizational performance constructs and other constructs (e.g., independent variables) are extracted through the same package of machine reading algorithm developed by \citet{chen2021machine} and annotated manually to three levels of categories capturing different approaches of defining organizational performance. At the most aggregate level (Level 0), constructs were grouped into performance and non-performance constructs. The performance category was further divided into Level 1- stakeholder relevance (investors, customers, employees, and the society) and the Level 2 subcategories int terms the specific approaches that inform measurement for each of the four stakeholders. This work seeks to train NLP and ML models to automatically classify the labels into Level 0, Level 1, and Level 2 categories given the constructs. 
Consider the sentence the following sentence: 
\begin{quote}
    \textit{A change in customer satisfaction will be correlated with a change in financial performance, a time series analysis was conducted.}
\end{quote}  

The constructs extracted from this sentence would be \textit{change in customer satisfaction} and \textit{change in financial performance}. Since \textit{customer satisfaction} and \textit{financial performance} capture an organization's effectiveness in the creation and delivery of well-being for its stakeholders (customers and investors, respectively), it will be treated as an instance of the \textsc{Performance} class. 


\section{Related work} \label{related work}
To the best of our knowledge, there is no prior work specifically addressing classification of performance and non-performance labels (i.e., Level 0).
\citet{10.1287/mksc.1110.0700} take a supervised learning approach to estimating readability and subjectivity. They train a 4-gram dynamic language model to derive the probability of subjectivity in customers' review. 

\citet{li2020theoryon} propose behavioral ontology learning from text based on which they develop a search engine which will help the researchers to search for the constructs, construct relationships, antecedents and consequents. They are automatically extracting the hypotheses, constructs, relation and network. Then they design TheoryOn, which utilizes sentence classifier to extract hypothesis using rule based and deep learning methods. This present research has a similar purpose, but our approach engages human domain experts on organizational performance in the beginning, and uses supervised learning to train classification models. In addition, we append our limited training data with context embedding by drawing surrounding sentences from all publications where our constructs were extracted and S\&P company reports where performance constructs are often discussed. In this way, we create more features to predict the classification, and the results improved in disaggregate-level classification, when the training data is very limited.

For comparing and analyzing the causal theories, \citet{mueller2015meta} suggests a new method. It represents the causal theories with a meta-model and also gives a taxonomy of inter-theory relationships. The proposed approach visualizes the evolution of theories, and identifies the contributions of a theory in a network of theories. 
It also creates a new ones by identifying the differences and similarities of related theories easily.

To the existing meta-model of causal theories, \citet{mueller2016theory} presents a meta-model of empirical study, methods for automatically inferring the expected empirical outcome of a causal theory, and proposes a visualization of relationship between theory and empirical data.

\citet{ludwig2020using} uses natural language processing techniques to tackle the construct identity problem. The work compares a GloVe word embedding model and different document projection methods with latent semantic analysis (LSA). The comparison of the results of GloVe and LSA show that both of them result in almost equal performance.

\citet{larsen2016tool} designed and evaluated the construct identity detector (i.e., constructs which share semantically similar measurement items), by adapting Natural language algorithms.The recall performance in full text results show that the participants can find relevant articles containing specific construct on average only 9\% of the time and 3\% of the time participants can find the relevant articles containing a pair of common constructs.

\section{Approach} \label{sec:approach}
To ground the NLP models into the domain of organizational research, we first collected a sample of papers related to organizational research in social sciences. We restricted our search of papers based on the explicit inclusion of organizational performance as part of the research question. In line with the new paradigm of multi-stakeholder and multi-dimensional conceptualization of corporate purpose \citep{harrison20202019}, we defined organizational performance as an organization’s effectiveness in meeting the expectations of two or more stakeholder groups (investors, employees, customers, and communities).
 
Based on the ISI Web of Science database of publications, all empirical publications (excluding meta-analysis) with at least one keyword directly suggesting a stakeholder group\footnote{The keywords indicating stakeholders were stakeholder*, investor*, shareholder*, owner*, and financ* for investors; customer*, consumer*, and user* for consumers; employee*, worker*, workforce*, labor*, labour*, and human resource* for employees; and communit*, societ*, environment*, climate*, natural resource*, responsib*, and social performance* for the community.} were retrieved and preprocessed. 
A snowball approach was adopted, in which each newly found performance construct was added as a new keyword for the next search until no new construct was found. With the pool of papers collected above, we further shortlisted papers that included theoretical developments related to performance measures concerning at least two stakeholder groups. This sample represents high-quality scientific journal articles and offers a viable corpus of testable knowledge (i.e., hypotheses) concerning organizational performance.
 
The primary studies included two stakeholder groups for measuring organizational performance: the correlations between a factor and at least two stakeholder values. In total, we have identified and downloaded 138 peer-reviewed articles published between 1990 and 2018. We further removed 13 papers of which the PDFs were of poor quality for optical character recognition (OCR). The remaining 125 papers represent cross-disciplinary literature in social sciences in 1990-2018 to explain different organizational performance dimensions.
 
Five doctoral students were engaged to manually extract all the hypotheses, and then deconstruct them into separate constructs.

We follow the taxonomy on organizational performance developed by \citet{chen2021}. The details about the labels and the number of examples per class present in our data is given in Table~\ref{tab:taxonomy}: 

Specifically, they define investor benefits (INV) as organizational performance concerning investors (e.g., shareholders) or, more specifically, the economic outcome of the firm, which benefits investors. They distinguish financial performance based on its factual basis, time horizon, as well as subjectivity. Specifically, to capture the factual basis of past performance and future expectations, they first distinguish (a) accounting-based performance (INV1) (e.g., return on assets, profitability, labor productivity, asset efficiency) and (b) stock market performance (INV2) (e.g., net income over the market value of equity, Tobin’s q, market-to-book value). To capture an organization’s long-term viability that is not often captured by accounting or stock market measures, we then include (c) growth of the firm (INV3) (e.g., sales growth, profit growth). To capture subjectively measured performance, they further include (d) survey-based satisfaction for financial positions of the firm (INV4) (e.g., a manager’s reporting, and an expert’s assessment).  \\

\begin{table}[ht!]
\resizebox{\columnwidth}{!}{%
        \begin{tabular}{llll}
        \toprule
          \textbf{Level} & \textbf{ Subcategory} & \textbf{Examples per class}   \\
        \midrule
        Level 0 & Performance  & 268  \\
               & Non-Performance & 421 \\
        \midrule
        Level 1 & Investors  & 87  \\
               & Customers  & 43 \\
               & Employees  & 55 \\
               & Society  &  83 \\
        \midrule
        Level 2 & Accounting-based performance (INV1) & 60 \\
         & Stock market-based performance (INV2) & 2 \\
         & Survey-based performance (INV3) & 11 \\
         & Growth-based performance (INV4) & 5 \\
         & Customer commitment (CUS1) & 4 \\
         & Customer satisfaction (CUS2) & 4 \\
         & Brand recognition and reputation (CUS3) & 17\\
         & Product \& service quality (CUS4) & 18\\
         & Employee commitment (EMP1) & 28\\
         & Employee satisfaction (EMP2) & 4 \\
         & Employee compensation, protection and benefits (EMP3) & 23 \\
         & Employee health (EMP4) & 0 \\
         & Symbolic socially responsible efforts (SOC1) & 6 \\
         & Substantive social impact and performance (SOC2) & 0 \\
         & Symbolic environmentally responsible efforts (SOC3) & 39 \\
         & Substantive environmental impact and performance (SOC4) & 16 \\
         & Uncategorized or combined (SOC5) & 22 \\

        \end{tabular}
        }
        \caption{Taxonomy of the organizational performance and the number of examples per class present in our dataset. Level 1 and Level 2 are the subcategories of the performance label in Level 0.}
        \label{tab:taxonomy}
        \end{table}
 Second, \citet{chen2021} define customer benefits (CUS) as organizational performance for customers or, more specifically, the benefit and the utility of products$/$services the firm creates for and delivers to customers. From a marketing perspective, performance is the extent to which an organization has satisfied its customers  \citet{neely1999performance,andy1995performance}. They include four different aspects of customers’ perspectives: (a) customer commitment (CUS1) (e.g., customer loyalty, customer retention), (b) customer satisfaction (CUS2) (e.g., satisfaction with product/service quality), (c) customer recognition of the firm (CUS3) (e.g., public reputation), and (d) objectively measured product/service quality (CUS4) (e.g., new product innovation, product safety). 
 
Third, they define employee benefits (EMP) as organizational performance for employees or, more specifically, the benefits and welfare employees receive from an organization. Following  \citet{clarkson2016stakeholder} , they include (a) employee commitment (EMP1) (e.g., turnover, organizational commitment), (b) employee satisfaction (EMP2) (e.g., job satisfaction, perceived justice), (c) employee compensation, protection, and benefits (EMP3) (e.g., compensation, job security), and (d) employee health (EMP4) (e.g., job burnout, physical health indicators). 
 
Community/environment benefits (SOC) are defined as organizational performance for the community/environment or, more specifically, an organization’s efforts and impacts on addressing societal, environmental, and general public concerns. Community/environment benefits can be distinguished into five subcategories including (a) symbolic measures of social concerns (SOC1) (e.g., societal mission statement, meeting agenda on societal issues), (b) substantive impact on the community (SOC2) (e.g., donation and philanthropy), (c) symbolic measures of environmental concerns (SOC3) (e.g., environmental mission statement, meeting agenda on environmental issues), (d) substantive impact on the natural environment (SOC4) (e.g., pollution control, waste disposal); and (e) combined index (SOC5) (e.g., the quality of social and environmental reporting). 
 
Finally, we added an additional category of performance to their taxonomy for constructs that are unclear on which subgroup it may belong to. We label them as unclassified constructs. Constructs that do not belong to any of the above categories will be labeled as non-performance.
 
Five PhD students in organization science, finance, and business were trained to manually label the data. Each construct is labeled independently by at least two coders. The initial inter-rater agreement was 81\%. After a direct conversation about disagreements, the inter-rater agreement reached 100\%. 
 

After extracting the constructs from the research documents, we have a total of 689 constructs. 
We split the data using stratified 5-fold cross validation to train and evaluate a series of models using BERT, multinomial naïve bayes, logistic regression, XGBoost, SVM, 
and fasttext using both latent and engineered features.
We report our results in section~\ref{results ans discussion}. 

\begin{table}[ht!]
\resizebox{\columnwidth}{!}{%
        \begin{tabular}{llll}
        \toprule
          & \textbf{Feature} & \textbf{ Description}   \\
        \midrule
         & Bag of $n$-gram values & [1,3]  \\
        Engineered features & seed & [1,4] \\
        
        \end{tabular}
}
        \caption{Counts with bag of $n$-gram for the engineered features}
        \label{tab:engineeredfeatures-example}
        \end{table}



\section{Results and discussion} \label{results ans discussion}
Performance for each of the machine learning models described in Section~\ref{sec:approach} are summarized in Tables~\ref{tab:performance}, \ref{tab:level-1}, \ref{tab:level-2}. 
For engineered features, we simply use TF-IDF scores and raw counts for token $n$-grams in the range of 1-3.
The results show that a simple logistic regression using counts and TF-IDF scores to represent token $n$-gram features outperforms all other models for Levels 0 and 1, while the same linear model with embedding-based features work best for the larger number of categories in the Level 2 classification task. 
\begin{table}[ht!]
\resizebox{\columnwidth}{!}{%
\begin{tabular}{l|l|llll}
\toprule
\multicolumn{1}{c}{} & \multicolumn{1}{l}{Model} & \multicolumn{1}{l}{P} & R & F1  \\ 
\midrule

 & XGBoost & 0.74 & 0.71 & 0.72 \\
Engineered features & Multinomial naïve Bayes & 0.80 & 0.81 & 0.80 \\
 & SVM & 0.81 & 0.81 & 0.81 \\
& \bestres Logistic regression & \bestres 0.81 & \bestres 0.81 & \bestres 0.81 \\
\midrule
 & FastText & 0.84 & 0.79 & 0.81 \\
 & Fine-tuned BERT & 0.59 & 0.51 & 0.40 \\
Latent features & SpaCy embeddings with LR & 0.74 & 0.75 & 0.75 \\
 & Elmo embeddings with LR & 0.78 & 0.78 & 0.78 \\
 
\end{tabular}
}
\caption{Classifier performance for the Level 0 label. P, R \& F1 are given in terms of macro average which is obtained by averaging the results across seed values 1 to 4.}
\label{tab:performance}
\end{table}

\begin{table}[ht!]
\resizebox{\columnwidth}{!}{%
\begin{tabular}{l|l|llll}
\toprule
\multicolumn{1}{c}{} & \multicolumn{1}{l}{Model} & \multicolumn{1}{l}{P} & R & F1  \\ 
\midrule
 
 & XGBoost &0.75 &0.61 &0.66 \\
Engineered features & Multinomial naïve Bayes & 0.69 & 0.72 & 0.70 \\
 & SVM & 0.77 & 0.72 & 0.74 \\
 & \bestres Logistic regression & \bestres 0.73 & \bestres 0.79 & \bestres 0.75 \\
\midrule
 & FastText &0.74 &0.29 &0.26 \\
 & Fine-tuned BERT & 0.92 & 0.20 & 0.15 \\
Latent features & SpaCy embeddings with LR & 0.68 & 0.78 & 0.72 \\
 & Elmo embeddings with LR & 0.72 & 0.71 & 0.71 \\
\end{tabular}
}
\caption{Classifier performance for the Level 1 label. P, R \& F1 are given in terms of macro average which is obtained by averaging the results across seed values 1 to 4.}
\label{tab:level-1}
\end{table}

\begin{table}[ht!]
\resizebox{\columnwidth}{!}{%
\begin{tabular}{l|l|llll}
\toprule
\multicolumn{1}{c}{} & \multicolumn{1}{l}{Model} & \multicolumn{1}{l}{P} & R & F1  \\ 
\midrule

 & XGBoost & 0.74 & 0.35 & 0.39 \\
Engineered features & Multinomial Naïve Bayes & 0.56 & 0.54 & 0.52 \\
 & SVM & 0.68 & 0.54 & 0.57 \\
 & Logistic regression & 0.61 & 0.59 & 0.58 \\
\midrule
 & FastText & 0.93 & 0.08 & 0.06 \\
 & Fine-tuned BERT & 0.98 & 0.06 & 0.05 \\
Latent features & \bestres SpaCy embeddings with LR & \bestres 0.59 & \bestres 0.65 & \bestres 0.60 \\
 & Elmo embeddings with LR & 0.59 & 0.59 & 0.58 \\
\end{tabular}
}
\caption{Classifier performance for the Level 2 label. P, R \& F1 are given in terms of macro average which is obtained by averaging the results across seed values 1 to 4.}
\label{tab:level-2}
\end{table}




\section{Conclusion} \label{conclusion}

In this work, we compare approaches for aligning concepts to different levels of a taxonomy. For small label sets (Levels 0 and 1), we found that a simple logistic regression classifier using TF-IDF scores of $n$-grams as features achieves the best overall performance, far exceeding the recall and F1 achieved by fine-tuned transformers.  
For our most granular label set (Level 2), word embeddings are more informative features overall.
In future work, we seek to explore few shot learning approaches to classifying rare taxonomic categories.



\bibliographystyle{acl_natbib}

\bibliography{main}

\begin{thebibliography}{13}
\expandafter\ifx\csname natexlab\endcsname\relax\def\natexlab#1{#1}\fi

\bibitem[{Andy(1999)}]{neely1999performance}
Neely Andy. 1999.
\newblock The performance measurement revolution: why now and what next?
\newblock \emph{International journal of operations \& production management},
  19(2):205--228.

\bibitem[{Andy et~al.(1995)Andy, Mike, and Ken}]{andy1995performance}
Neely Andy, Gregory Mike, and Platts Ken. 1995.
\newblock Performance measurement system design. a literature review
  andresearch agenda.
\newblock \emph{International journal of operations \& production management},
  15(4):80--116.

\bibitem[{Chen et~al.(2021{\natexlab{a}})Chen, Montano-Campos, Zadrozny, and
  Canfield}]{chen2021machine}
Victor~Zitian Chen, Felipe Montano-Campos, Wlodek Zadrozny, and Evan Canfield.
  2021{\natexlab{a}}.
\newblock Machine reading of hypotheses for organizational research reviews and
  pre-trained models via r shiny app for non-programmers.
\newblock \emph{arXiv preprint arXiv:2106.16102}.

\bibitem[{Chen et~al.(2021{\natexlab{b}})Chen, Zhong, Duran, and
  Sauerwald}]{chen2021}
Victor~Zitian Chen, Meng Zhong, Patricio Duran, and Steve Sauerwald.
  2021{\natexlab{b}}.
\newblock Multi-stakeholder benefits: A meta-analysis of different theories.
\newblock \emph{working paper}.

\bibitem[{Clarkson(2016)}]{clarkson2016stakeholder}
Max~BE Clarkson. 2016.
\newblock \emph{A stakeholder framework for analysing and evaluating corporate
  social performance}.
\newblock University of Toronto Press.

\bibitem[{Ghose et~al.(2012)Ghose, Ipeirotis, and Li}]{10.1287/mksc.1110.0700}
Anindya Ghose, Panagiotis~G. Ipeirotis, and Beibei Li. 2012.
\newblock \href {https://doi.org/10.1287/mksc.1110.0700} {Designing ranking
  systems for hotels on travel search engines by mining user-generated and
  crowdsourced content}.
\newblock \emph{Marketing Science}, 31(3):493–520.

\bibitem[{Harrison et~al.(2020)Harrison, Phillips, and
  Freeman}]{harrison20202019}
Jeffrey~S Harrison, Robert~A Phillips, and R~Edward Freeman. 2020.
\newblock On the 2019 business roundtable “statement on the purpose of a
  corporation”.
\newblock \emph{Journal of Management}, 46(7):1223--1237.

\bibitem[{Kang et~al.(2020)Kang, Cai, Tan, Huang, and Liu}]{kang2020natural}
Yue Kang, Zhao Cai, Chee-Wee Tan, Qian Huang, and Hefu Liu. 2020.
\newblock Natural language processing (nlp) in management research: A
  literature review.
\newblock \emph{Journal of Management Analytics}, 7(2):139--172.

\bibitem[{Larsen and Bong(2016)}]{larsen2016tool}
Kai~R Larsen and Chih~How Bong. 2016.
\newblock A tool for addressing construct identity in literature reviews and
  meta-analyses.
\newblock \emph{MIS Q.}, 40(3):529--551.

\bibitem[{Li et~al.(2020)Li, Larsen, and Abbasi}]{li2020theoryon}
Jingjing Li, Kai Larsen, and Ahmed Abbasi. 2020.
\newblock Theoryon: A design framework and system for unlocking behavioral
  knowledge through ontology learning.
\newblock \emph{MIS Quarterly}, 44(4):1733--1772.

\bibitem[{Ludwig et~al.(2020)Ludwig, Funk, and Mueller}]{ludwig2020using}
Siegfried Ludwig, Burkhardt Funk, and Benjamin Mueller. 2020.
\newblock Using natural language processing techniques to tackle the construct
  identity problem in information systems research.

\bibitem[{Mueller(2015)}]{mueller2015meta}
Roland~M Mueller. 2015.
\newblock A meta-model for inferring inter-theory relationships of causal
  theories.
\newblock In \emph{2015 48th Hawaii International Conference on System
  Sciences}, pages 4908--4917. IEEE.

\bibitem[{Mueller(2016)}]{mueller2016theory}
Roland~M Mueller. 2016.
\newblock Theory-data maps: a meta-model and methods for inferring and
  visualizing relationships between causal theories and empirical evidences.
\newblock In \emph{2016 49th Hawaii international conference on system sciences
  (HICSS)}, pages 5288--5297. IEEE.

\end{thebibliography}


\end{document}